**In Love With a Robot: the Dawn of Machine-To-Machine Marketing**

By E.Kotomin

A short while ago I got my heart broken by a robot. Lost on a popular website I was offered direct chat assistance. The brief dialogue with a young woman named Lisa was highly entertaining. She expertly answered all my questions, we shared a few jokes and just as I was about to ask her out, it became obvious that as much as I had enjoyed our brief time together, we weren't meant to be - Lisa was but a line of code in cyberspace.

If you haven't been paying attention to the news lately, artificial intelligence is on the rise. Robots manage 70% of the US stock trading volume[1], academics consider intelligent software assistance one of 10 emerging technologies[2], and if there ever was a company capable of bringing the concept to the masses, it is certainly Apple, who, coincidently, made it the key marketing point of the latest iPhone.

Considering the pace of technological progress, it is safe to assume that in another decade virtual assistants are going to evolve into highly sophisticated cloud programs. Accessible through a range of devices anytime anywhere, and hooked up to the Social Media mainframe, my personal robot, let's call him Floyd, will know everything there is to know about me- my preferences in food, fashion and people, my financial state, hobbies and holiday plans. Utilising petabytes of data, he will be able to make autonomous decisions on my behalf. Floyd will know which configuration of iPhone 15 I would like, how much I can afford to spend on it and whether I would prefer to wait a week and get it cheaper or pay the premium and have it delivered to the office tomorrow.

On the vendor side, my digital PA will be confronted with equally advanced robots that will understand anthropomorphic patterns, recognising the difference between the CEO of GE and me but also between Tom Hanks and Keith Richards- a distinction based not only on income and status but on an exhaustive social profile. The impact autonomous robotic retail is going to have on our lives will be of a scale comparable to the invention of a battery. With both parties providing services with digital precision around the clock, the concepts of store opening hours and busy hotlines will become history. When I am able to order a pair of Converse shoes while watching iRobot by simply picking up the phone and saying: "get me one of those", the return on investment in product placement will become more transparent. Linked to smart TV sets, Floyd will understand what I mean and know what size to order.

---

[1] E. Kaufmann, C.Levin, 'Preventing the Next Flash Crash', in The New York Times. 5 May 2011, viewed on 17 February 2013 <http://www.nytimes.com/2011/05/06/opinion/06kaufman.html?_r=0>
[2] E. Naone, 'TR10: Intelligent Software Assistant', MIT Technology Review. March/April 2009, viewed on 17 February 2013 <http://www2.technologyreview.com/article/412191/tr10-intelligent-software-assistant/>

Some of the challenges of that digital paradise are fairly predictable as well. Firstly, and most obviously: security. With my banking details in his pocket, my virtual assistant is going to need protection. Secondly, with the concept of personal gifts degrading to a transaction, life will become less emotional. Although I will never forget another anniversary, my girlfriend will know that I didn't spend sleepless nights deciding what to get her and days of agony trying to acquire it – I simply asked Floyd to pick something up based on her profile. If you follow the concept through, it will lead to a rather twisted place - with my digital assistant picking up perfect gifts and compiling personal notes, who is the woman romantically involved in really – me or him?

We invest a lot in artificial intelligence. Whether personally, by buying the latest cell phone; or corporately, by building advanced infrastructure that we expect to excel at trading stocks, predicting the weather or thwarting security threats. Robots do not need breaks and can analyse exorbitant amounts of data billions of times faster than their creators. But therein lies the question: will artificial intelligence become truly intelligent or will it always be restricted to analysing facts in the same manner a pocket calculator is limited to a set of predetermined actions with a limited number of figures?

For marketeers, the dichotomy suggests two scenarios both presuming cardinal changes to the industry. If, pumped up with data, Floyd will evolve not just to guess the colour of the tie to buy but to **understand** the abstract concept of (my) taste, he will be able to make autonomous decisions and thus will become the legitimate, if not primary, target for marketing. Considering how fast we have moved from monochrome mobiles to dual-core processor smartphones, this futuristic vision may become reality much faster than we anticipate.

Even if virtual assistants are really just statisticians with a mind limited by the horizon of a users' shopping history, cloud robotics is going to change the essence of commerce the fundamentals of which have been unaltered for thousands of years. With no fear, hunger, requirement for social recognition, coupled with an emotional detachment, Floyd will evaluate offers based on a set of tangible factors- size, quality, ingredients, price and so on. With that in mind, what will Machine-To-Machine marketing appeal to? And more importantly, who is going to be shaping that industry - myself, a bunch of IT engineers or robots themselves? Considering they will be the ones doing most of the shopping and the ever- growing complexity of the code, who is going to understand the new generation of consumers better?

The Internet is really a home for Floyd and his kin. The crash of 2.45, the most prominent recent example of technology going awry, clearly demonstrates the extent to which we have lost control over cloud assistants. On May 6 2010, with no human interference, a trader robot caused the Dow Jones to drop 700 points in just a few minutes, costing an estimated $1 trillion in market

capitalisation. The event was not the first of its kind and the market bounced back but in the words of the CFTC commissioner Scott O'Malia, "Do we treat rogue algorithms like rogue traders?[3]"

The long-term challenges of autonomous cloud robotics will fall on the shoulders of later generations. In the world of today, the robotic marketplace is already very real and rapidly growing, moving from black- box trading to grocery shopping. Driven by the technology race, robots are getting faster and omnipresent; their code - increasingly more convoluted. Whether or not machines will ever understand **us,** considering they are being groomed to take over most of the world's infrastructure, from economy to security, we need to understand **them.** Otherwise, when the lights go out we will not even be able to go back to candles – based on my shopping history Floyd will not have bought them.

---

[3] U.S. Commodity Futures Trading Commission, 'Press Release PR5920-10', 7 October 2010, viewed on 17 February 2013,<http://www.cftc.gov/PressRoom/PressReleases/pr5920-10>